%% file: main.tex
\documentclass[10pt,twocolumn,letterpaper]{article}

\usepackage[pagenumbers]{cvpr}      %

\usepackage{booktabs}
\usepackage{multirow}
\usepackage{graphicx}
\usepackage{array}
\usepackage{xcolor}
\usepackage{colortbl}
\usepackage{pifont}
\usepackage{tabularx}
\usepackage{makecell}

\definecolor{cvprblue}{rgb}{0.21,0.49,0.74}

\usepackage[pagebackref,breaklinks,colorlinks,allcolors=cvprblue]{hyperref}

\def\confName{CVPR}
\def\confYear{2027}

\title{Beyond Additive Composition: Interventional Modeling of State-Dependent Cross-Degradation Responses for Composite Image Restoration}

\author{
Zongda Tang$^{1}$ \quad Jie Feng$^{1}$\thanks{Corresponding author.} \quad Wenda Zhao$^{2}$ \quad Mingtao Feng$^{1}$\\
Yuping Liang$^{1}$ \quad Ronghua Shang$^{1}$ \quad Weisheng Dong$^{1}$\\
$^{1}$Xidian University \qquad $^{2}$Dalian University of Technology
}

\begin{document}
\maketitle
\input{sec/0_abstract}

\input{sec/teaser}
\input{sec/1_intro}
\input{sec/2_related_work}

\input{sec/3_method}

\input{sec/4_expriments}

\input{sec/5_conclusion}

\input{sec/LLM_Usage_Statement}
{
    \small
    \bibliographystyle{ieeenat_fullname}
    \bibliography{main}
}

\end{document}

%% file: sec/0_abstract.tex
\begin{abstract}
Composite image restoration aims to recover images corrupted by multiple coexisting degradations. Existing methods predominantly decompose composite corruptions into individual factors, relying on these isolated priors to guide restoration. However, coexisting degradations are inherently interdependent: the manifestation of one degradation dynamically shifts with the states of others, inducing non-additive interactions that factor-wise paradigms fail to capture. To bridge this gap, we formulate state-dependent cross-degradation responses and propose PROBE, an interventional framework that explicitly models how coexisting degradations influence one another. Specifically, we first construct unified degradation axes by aligning visual and textual degraded-clean offsets. Along these axes, a calibrated coordinate head estimates current degradation states, while bidirectional feature perturbations are injected to probe cross-coordinate responses under the current states. Contrasting these responses with corresponding intensity-matched single-degradation references quantifies intrinsic non-additive deviations, allowing the resulting response descriptors and state coordinates to jointly guide restoration. For comprehensive evaluation, we introduce SCD-15, a new benchmark  covering all 15 nonempty combinations of four degradation factors at three severity levels. Experiments demonstrate that PROBE achieves state-of-the-art performance on the CDD-11, AWMM-100K, and SCD-15 composite restoration datasets using a compact model with modest GPU memory requirements.
\end{abstract}

%% file: sec/teaser.tex
\begin{figure*}[!t]
    \centering
    
    \includegraphics[width=\textwidth]{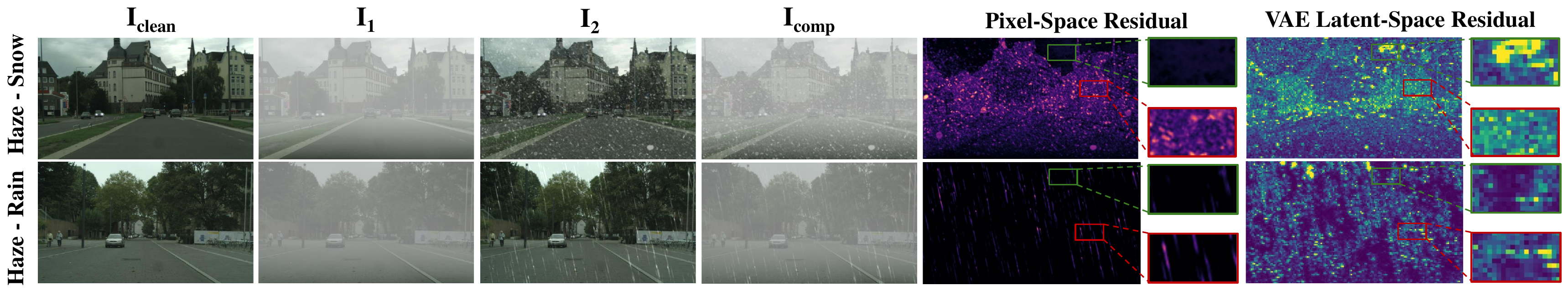}
    \vspace{-2em}
    \caption{Non-additive residuals in pixel space and VAE latent space on SCD-15. Each row shows the clean image $I_{\mathrm{clean}}$, the haze-only image $I_1$, the snow-only (top) or rain-only (bottom) image $I_2$, and the composite image $I_{\mathrm{comp}}$. The last two columns show the magnitude of $E_\Psi=\Psi(I_{\mathrm{comp}})-[\Psi(I_1)+\Psi(I_2)-\Psi(I_{\mathrm{clean}})]$, with $\Psi$ denoting the identity map in pixel space or the scaled posterior mean from a pretrained VAE encoder in latent space. The latent heatmaps show the root-mean-square residual across channels at each spatial location.}
    \label{fig:motivation1}
    \vspace{-1em}
\end{figure*}

%% file: sec/1_intro.tex
\section{Introduction}
\label{sec:intro}

High-quality imaging is essential for reliable visual perception, yet adverse conditions can simultaneously introduce multiple degradations, such as low illumination, haze, and rain, which severely compromise image quality. Composite image restoration seeks to recover clean images from such observations~\cite{OneRestore}, accounting for both the constituent degradations and their interactions within the image formation process.

All-in-one image restoration handles diverse degradations within a shared model using degradation-aware representations and adaptive conditioning~\cite{AirNet,PromptIR,MoCEIR}. Recent methods extend this paradigm to composite restoration through scene-level and factor-level descriptions~\cite{OneRestore,DPL,AllRestorer}. While effective in characterizing degradation content and composition, these approaches implicitly treat coexisting degradations as additive or separable, overlooking their complex interactions in real-world imaging. 

To examine whether the effects of two degradations are additive, we study matched haze--snow and haze--rain examples from SCD-15, which models coupled composite corruptions following OneRestore~\cite{OneRestore}. For each example in Fig.~\ref{fig:motivation1}, we sum the changes induced by the two single degradations relative to the same clean image and compare this prediction with the change induced by the composite degradation. In pixel space, the prediction leaves structured residuals, with snow-like patterns in the haze--snow example and rain streaks in the haze--rain example. We repeat the comparison in the latent space of the Stable Diffusion v2.1 base VAE~\cite{SD2.1}, taking the scaled posterior mean as the representation of each image. The root-mean-square prediction error reaches 33.0\% of the observed composite latent shift for haze--snow and 35.7\% for haze--rain, with both shifts measured in root-mean-square magnitude. These substantial departures from additivity across both pixel and latent spaces motivate explicit modeling of dependencies between degradation factors alongside their individual states.

This non-additive behavior calls for examining how degradation effects vary with the states of other factors. Related adverse-weather studies recognize such coupling: heavy-rain restoration jointly considers rain streaks and atmospheric accumulation coupled through image formation~\cite{HeavyRain}, while RMR addresses coupled rain and mist by integrating single-degradation priors through bilevel optimization~\cite{RMR}. 
However, these methods are tailored to specific degradation pairs and only implicitly accommodate coupling during restoration. 
For general composite corruptions, merely estimating static degradation states is insufficient.  Effective restoration further requires capturing how shifting one factor dynamically alters the others. This motivates us to go beyond static state descriptions and explicitly formulate state-dependent cross-degradation responses as a complementary prior to guide restoration.

To this end, we propose PROBE, an interventional framework that characterizes composite degradations through both their states and responses to controlled interventions. We align textual degraded-clean offsets with paired visual residuals to establish semantically grounded degradation axes in the learned feature space, providing a common basis for coordinate estimation and local dependency probing. A calibrated coordinate head  describes the current degradation state, while bidirectional perturbations along each axis measure the responses of other coordinates. Contrasting these responses against  intensity-matched single-degradation references allows us to distinguish composite-specific non-additive effects from cross-coordinate sensitivity already induced when probing isolated degradations. The degradation coordinates and reference-relative response deviations jointly guide a NAFNet-based baseline through multi-scale modulation~\cite{NAFNet}, enabling adaptive restoration informed by state-dependent degradation interactions.

Our contributions can be summarized as follows:
\begin{itemize}
    \item We reveal that composite degradations inherently violate additive separability in both pixel and latent spaces. Moving beyond static factor-wise paradigms, we introduce state-dependent cross-degradation responses to explicitly characterize how coexisting degradations dynamically influence one another under specific states.
    \item We propose PROBE, an interventional framework that provides a principled formulation and computation mechanism for these responses. By injecting bidirectional perturbations along semantically anchored degradation axes and contrasting against single-degradation references, PROBE accurately quantifies non-additive interactions to guide restoration.
    \item PROBE achieves state-of-the-art performance on CDD-11, AWMM-100K, and our new SCD-15 benchmark with modest model complexity. SCD-15 incorporates three degradation severity levels and evaluates generalization to degradation combinations unseen during training.
\end{itemize}

%% file: sec/2_related_work.tex
\section{Related Work}
\label{sec:formatting}

\subsection{All-in-One Image Restoration}

All-in-one image restoration addresses multiple degradation types and severity levels within a unified model. Early studies explore noise-level conditioning and shared restoration pretraining~\cite{FFDNet,IPT}, followed by degradation-aware representations based on contrastive learning, ingredient integration, prompting, and frequency modulation~\cite{AirNet,IDR,PromptIR,AdaIR,CPL}. Expert routing and degradation-prior retrieval adapt shared reconstruction to different inputs~\cite{MoCEIR,R2R}, while task uncertainty regularization addresses optimization imbalance across tasks~\cite{TUR}. Together with general restoration architectures~\cite{MPRNet,Uformer,Restormer,NAFNet}, these advances provide effective frameworks for unified restoration.

Language and generative priors also improve restoration by incorporating semantic degradation cues. DA-CLIP introduces vision--language representations sensitive to degradations, while InstructIR guides restoration with natural-language instructions~\cite{DA-CLIP,InstructIR}. Related studies explore degradation removal in textual representations~\cite{TextualDegRemoval}, and unfolding frameworks incorporate prompts and vision--language guidance into iterative reconstruction~\cite{UPRestorer,VLUNet}. Generative approaches complement these designs through degradation-conditioned latent diffusion~\cite{AutoDIR}, selective mappings, visual degradation instructions, and degradation-aware low-rank adaptation~\cite{DiffUIR,Defusion,DOD}. Despite their effectiveness, these methods primarily address different degradation types as separate restoration tasks within a unified framework. PROBE follows the growing interest in composite restoration, where multiple degradation factors coexist within a single image.

\subsection{Composite Image Restoration}

Composite image restoration addresses coexisting degradations within an image. Early work primarily targets specific combinations, including joint raindrop--streak removal~\cite{CCN} and hybrid-distortion restoration~\cite{FDRNet}, with subsequent studies extending to more general complex degradations~\cite{UniRes,FDI}. Recent methods characterize composite inputs through degradation-aware representations and adaptive reconstruction. OneRestore introduces a composite imaging model and scene descriptors~\cite{OneRestore}. DPL learns degradation- and factor-level prototypes, including shared prototypes motivated by degradation correlations~\cite{DPL}. AllRestorer combines visual--textual descriptions with adaptive degradation weights~\cite{AllRestorer}, while CART employs composition-aware task conditioning and adaptive expert routing~\cite{CART}. However, these methods primarily focus on static composition identification or factor-level correlations, leaving state-dependent non-additive interactions unexplored.

Coupled-degradation studies examine these dependencies more directly. Heavy-rain restoration models the coupled effects of rain streaks and atmospheric accumulation within a unified image formation process~\cite{HeavyRain}, while RMR integrates single-degradation priors through bilevel optimization for coupled rain and mist~\cite{RMR}. Whereas these methods design restoration strategies for specific couplings, 
PROBE offers a general interventional framework that explicitly probes cross-degradation responses across arbitrary combinations to guide restoration.

%% file: sec/3_method.tex
\section{Analysis of State-Dependent Degradation Responses}
\label{sec:analysis}

\begin{figure}[t]
    \centering
    \includegraphics[width=0.9\linewidth]{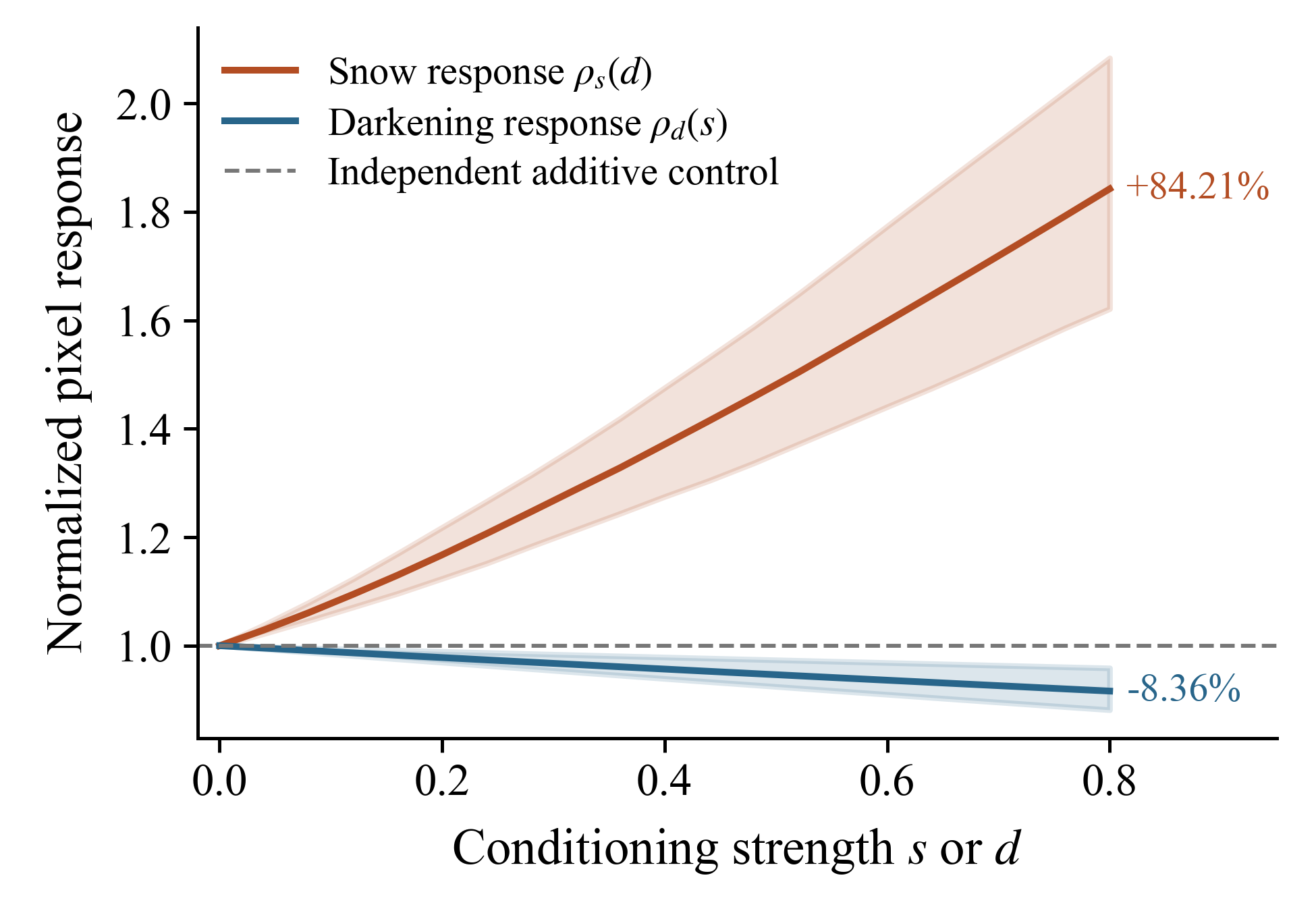}
    \vspace{-1em}
    \caption{State-dependent pixel responses in controlled low-light--snow synthesis. The probed factor increases from $0.3$ to $0.5$; the conditioning factor is fixed within pairs and swept over $[0,0.8]$ across pairs. The horizontal axis gives snow opacity for the blue darkening-response curve and darkening strength for the orange snow-response curve. RMS changes are normalized per scene by the response without the conditioning factor. Curves/bands show the median/interquartile range over 200 scenes; the dashed line is the independent additive control.}
    \label{fig:motivation2}
    \vspace{-1em}
\end{figure}

\begin{figure*}[t]
    \centering
    \includegraphics[width=\textwidth]{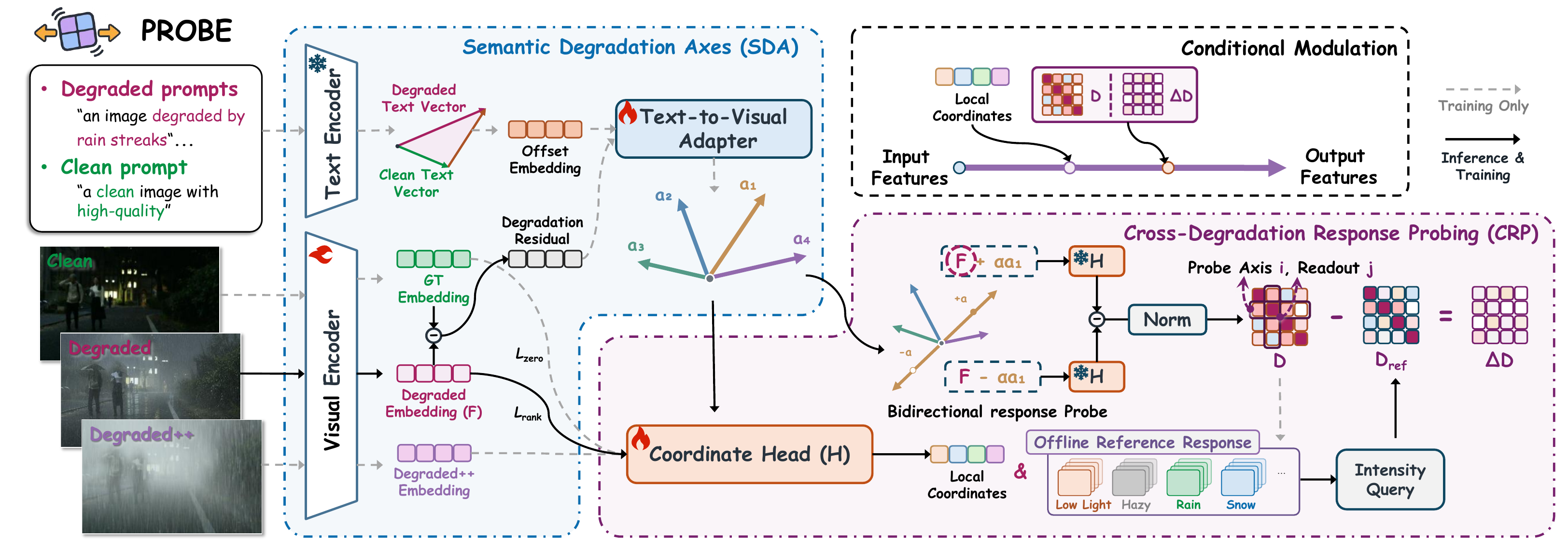}
    \vspace{-1.5em}
    \caption{Overview of PROBE. A NAFNet-based encoder-decoder is conditioned at the bottleneck and a decoder stage. Semantic degradation axes are derived from degraded-clean text offsets and aligned with paired visual residuals. An axis-conditioned head estimates local coordinates, while bidirectional axis-wise probing measures cross-coordinate responses. Current coordinates query an offline reference from single-degradation training samples to obtain response deviations. Coordinates and response descriptors guide reconstruction through conditional modulation. Dashed/solid arrows denote training-only/training-and-inference paths.}
    \label{fig:pipeline}
    \vspace{-1em}
\end{figure*}

While Fig.~\ref{fig:motivation1} reveals  departures from additivity, it does not explicitly show how the sensitivity to one degradation shifts with the severity of coexisting factors. To systematically probe this state dependence, we conduct a controlled interventional study on low-light--snow synthesis with 200 clean CDD-11 test images and Snow100K masks. Let $I_{d,s}$  denote  the synthesized image corrupted by darkening strength $d$ and snow opacity $s$ from a clean reference  $I_{\mathrm{clean}}$.  To measure the marginal effect of each factor under varying coexisting states, we apply a fixed severity increment from $0.3$ to $0.5$   to one degradation while sweeping the conditioning factor across  $[0,0.8]$:
\begin{equation}
\begin{aligned}
\Delta_d I(s) &= I_{d_1,s}-I_{d_0,s},\\
\Delta_s I(d) &= I_{d,s_1}-I_{d,s_0},
\end{aligned}
\label{eq:pixel_response}
\end{equation}
where $d_0=s_0=0.3$ and $d_1=s_1=0.5$ define darkening and snow increments at fixed $s$ and $d$, respectively.

For each scene, we measure response magnitude using the root mean square (RMS)~\cite{RMS} over all pixels and RGB channels. To quantify state dependence, we divide each conditional response magnitude by its value when the other degradation is absent:
\begin{equation}
\rho_d(s)=\frac{\|\Delta_d I(s)\|_{\mathrm{RMS}}}{\|\Delta_d I(0)\|_{\mathrm{RMS}}},
\;
\rho_s(d)=\frac{\|\Delta_s I(d)\|_{\mathrm{RMS}}}{\|\Delta_s I(0)\|_{\mathrm{RMS}}}.
\label{eq:normalized_pixel_responses}
\end{equation}

Under the conventional additive separability assumption, $I_{d,s}=I_{\mathrm{clean}}+U_d(I_{\mathrm{clean}};d)+U_s(I_{\mathrm{clean}};s)$, the marginal change induced by darkening or snow depends solely on its own severity. Consequently, $\Delta_d I(s)$ is invariant to $s$ and $\Delta_s I(d)$ to $d$, which strictly implies constant unit ratios $\rho_d(s)=\rho_s(d)\equiv 1$ across all conditioning states.

However, empirical observations in Fig.~\ref{fig:motivation2} starkly contradict this invariance: increasing snow opacity progressively suppresses the pixel response to a fixed darkening increment ($\rho_d(s) < 1$), whereas stronger darkening amplifies the response to a fixed snow increment ($\rho_s(d) > 1$).  This demonstrates that identical degradation increments elicit markedly different responses depending on the coexisting factor's state, with such cross-degradation coupling exhibiting clear asymmetry.  Since controlled image-pair interventions and isolated references ($\Delta I(0)$) are unavailable for a single degraded test image at inference time, these findings directly motivate our PROBE framework (Sec.~\ref{sec:method}), which emulates this interventional probing and reference contrasting in the learned feature space.

\section{Proposed Method}
\label{sec:method}

Motivated by the analysis in Sec.~\ref{sec:analysis}, we propose PROBE, an interventional restoration framework guided by degradation coordinates and cross-degradation response descriptors. We map an input image from pixel space into a learned representation space, where its degradation information is represented as a degradation state. Degradation coordinates quantify this state, with each coordinate indicating the relative severity of a specific degradation. The semantically anchored degradation axes (SDA) module aligns text-derived directions with visual degradation residuals. The cross-degradation response probing (CRP) module estimates coordinates along these axes with a calibrated head and probes them bidirectionally to measure cross-coordinate responses at the current state. Together, the coordinates and response descriptors characterize degradation severity and local inter-factor sensitivity, jointly conditioning restoration as illustrated in Fig.~\ref{fig:pipeline}.

\subsection{Semantically Anchored Degradation Axes}
\label{sec:axes}

We define a semantic probing direction for each of $K$ categories $\mathrm{deg}\in\mathcal{D}$. A visual encoder separate from the restoration baseline extracts full-resolution $C$-channel features from $I\in\mathbb{R}^{3\times H\times W}$. In each $16\times16$ spatial window $w$, pooled features and MLP-embedded local RGB statistics form $\mathbf{f}^w(I)\in\mathbb{R}^{C}$ for axis alignment, coordinate estimation, and probing. A clean image $I_{\mathrm{clean}}$ and its single-degradation counterpart $I_{\mathrm{deg}}$ yield the degradation-induced residual:
\begin{equation}
\mathbf{r}_{\mathrm{deg}}^w =
\mathbf{f}^w(I_{\mathrm{deg}}) -
\mathbf{f}^w(I_{\mathrm{clean}}).
\end{equation}

Using the pretrained CLIP text encoder $T$ from DA-CLIP~\cite{DA-CLIP}, a learnable adapter $g$ maps the embedding offset between degraded and clean prompts ($p_{\mathrm{deg}},p_{\mathrm{clean}}$)  to a semantic axis in the visual feature space:
\begin{equation}
\mathbf{a}_{\mathrm{deg}} =
\operatorname{Normalize}\left(
g(T(p_{\mathrm{deg}})-T(p_{\mathrm{clean}}))
\right),
\label{eq:axes}
\end{equation}
where $\mathbf{a}_{\mathrm{deg}}\in\mathbb{R}^{C}$ is $\ell_2$-normalized.  Enforcing cosine alignment between $\mathbf{a}_{\mathrm{deg}}$ and $\mathbf{r}_{\mathrm{deg}}^w$ grounds text-defined degradation semantics into the visual measurement space.

The resulting axis set $\mathbf{A}=\{\mathbf{a}_{\mathrm{deg}}\mid\mathrm{deg}\in\mathcal{D}\}$ provides a unified basis for estimating $K$ state coordinates and a $K\times K$ response matrix per window. Once learned, $\mathbf{A}$ remains fixed during restoration, requiring no text encoding at inference.

\subsection{Cross-Degradation Response Probing}
\label{sec:coordinate}
Using these axes, we establish a principled formulation and computation mechanism for state-dependent cross-degradation responses. Specifically, calibrated coordinates and bidirectional feature perturbations are combined to measure cross-coordinate responses in feature space, which are then contrasted with intensity-matched single-degradation references to isolate non-additive deviations.

\noindent\textbf{Coordinate calibration.}
The degradation axes specify probing directions but do not by themselves quantify degradation severity. To obtain a quantitative description of the current state, we calibrate a shared axis-conditioned head $h$ to estimate the relative severity of each degradation within a window:
\begin{equation}
\tau_{\mathrm{deg}}^w =
h(\mathbf{f}^w(I), \mathbf{a}_{\mathrm{deg}}),
\qquad
\boldsymbol{\tau}^w\in\mathbb{R}^{K}.
\label{eq:coordinates}
\end{equation}
Here, $\tau_{\mathrm{deg}}^w$ represents the relative severity of degradation $\mathrm{deg}$ in window $w$, and $\boldsymbol{\tau}^w$ collects the coordinates of all $K$ categories to quantitatively characterize the local degradation state. The head combines a scaled feature--axis projection with an additive MLP correction, sharing parameters across categories. Clean-image zero anchoring establishes a near-zero baseline, while severity ranking encourages coordinates to increase with degradation severity. The resulting coordinates encode relative degrees rather than probabilities or absolute physical parameters, providing a calibrated readout for subsequent response probing.

\noindent\textbf{Bidirectional response measurement.}
To capture how shifting one factor affects others under the current state, we perturb the window feature $\mathbf{f}^w(I)$ along each axis and measure the induced coordinate changes. For $i,j\in\{1,\ldots,K\}$, symmetric feature perturbations along $\mathbf{a}_i$ yield the finite-difference directional sensitivity:
\begin{equation}
R_{ij}^{w}=\frac{
h(\mathbf{f}^{w}(I)+\alpha\mathbf{a}_i,\mathbf{a}_j)
-h(\mathbf{f}^{w}(I)-\alpha\mathbf{a}_i,\mathbf{a}_j)
}{2\alpha}.
\label{eq:response}
\end{equation}
Here, $\alpha>0$ is the probing step. For small $\alpha$, this centered difference approximates the directional derivative of the $j$-th coordinate along $\mathbf{a}_i$ at the current window feature $\mathbf{f}^{w}(I)$. The response matrix $\mathbf{R}^{w}\in\mathbb{R}^{K\times K}$  measures local directional sensitivities across all factor pairs: rows index probing directions and columns index responding coordinates. Diagonal entries capture self-responses, while off-diagonal entries capture cross-coordinate responses without imposing symmetry.

To account for differences in response scale across coordinates, we normalize each column as $D_{ij}^{w}=R_{ij}^{w}/s_j$, where $s_j>0$ is a fixed RMS response scale estimated from single-degradation reference windows across all probing directions. The resulting $\mathbf{D}^{w}$ expresses each response relative to the typical sensitivity of its output coordinate.

\noindent\textbf{Single-degradation reference.}
Because all $K$ axes are probed for any input, cross-coordinate responses in $\mathbf{D}^{w}$ arise even under a single degradation.
To isolate composite-specific effects, we construct an offline severity-conditioned reference for each $\mathrm{deg}\in\mathcal{D}$ from training single-degradation windows.  Specifically, we partition the corresponding windows into eight quantile bins according to $\tau_{\mathrm{deg}}^w$. Each bin stores its mean coordinate and mean normalized response matrix. Linear interpolation between these mean coordinates defines $\overline{\mathbf{D}}_{\mathrm{deg}}(t)\in\mathbb{R}^{K\times K}$, which estimates the typical single-degradation response at coordinate $t$.

For an input window with estimated state $\boldsymbol{\tau}^w$, we query the intensity-matched references and subtract their state-weighted mixture from the measured response:
\begin{equation}
\raisebox{-6pt}{\(\displaystyle
\begin{aligned}
\quad \mathbf{D}_{\mathrm{ref}}^{w} &=
\sum_{\mathrm{deg}\in\mathcal{D}}
\omega_{\mathrm{deg}}(\boldsymbol{\tau}^{w})
\overline{\mathbf{D}}_{\mathrm{deg}}(\tau_{\mathrm{deg}}^{w}), \\
\Delta\mathbf{D}^{w} &= \mathbf{D}^{w} - \mathbf{D}_{\mathrm{ref}}^{w}.
\end{aligned}
\)}
\label{eq:reference}
\end{equation}
Mixture weights \(\omega_{\mathrm{deg}}(\boldsymbol{\tau}^w)\) are computed by subtracting each coordinate's clean baseline, rescaling by its single-degradation scale, and normalizing positive activations to sum to one. Each reference thus receives a nonnegative contribution per estimated degradation state, so \(\Delta\mathbf{D}^{w}\) explicitly quantifies non-additive response deviations from multi-degradation coupling.

\subsection{Coordinate--Response Conditioning}
\label{sec:conditioning}

A shared MLP $\phi$ encodes the degradation coordinates, normalized responses, and reference deviations of each window into a 16-dimensional restoration descriptor,
\begin{equation}
\mathbf{z}^{w}=\phi\!\left(
[\boldsymbol{\tau}^{w},\operatorname{vec}(\mathbf{D}^{w}),
\operatorname{vec}(\Delta\mathbf{D}^{w})]\right),
\label{eq:signature}
\end{equation}
where $\operatorname{vec}$ vectorizes each matrix. Spatially interpolating $\boldsymbol{\tau}^w$ and $\mathbf{z}^w$ yields the state map $\mathbf{T}\in\mathbb{R}^{K\times H\times W}$ and descriptor map $\mathbf{Z}\in\mathbb{R}^{16\times H\times W}$, respectively.

Following R2R~\cite{R2R}, our unconditioned baseline is a U-shaped encoder--decoder with NAFNet blocks~\cite{NAFNet}. At both the bottleneck and the quarter-resolution decoder stage, PROBE applies two modulation modules in sequence, conditioned first on $\mathcal{R}(\mathbf{T})$ and then on $\mathcal{R}(\mathbf{Z})$, where $\mathcal{R}$ resizes each map to the feature resolution. Each module modulates features $\mathbf{X}$ using condition $\mathbf{Q}$~\cite{FiLM,SFT,SPADE} according to
\begin{equation}
\mathcal{M}(\mathbf{X},\mathbf{Q})
=\mathbf{X}+\gamma(\mathbf{Q})\odot\operatorname{LN}(\mathbf{X})
+\beta(\mathbf{Q}),
\label{eq:modulation}
\end{equation}
where $\gamma$ and $\beta$ are learned convolutional scale and bias predictors, $\operatorname{LN}$ applies layer normalization across channels at each location, and $\odot$ denotes element-wise multiplication. Adding the predicted residual to the degraded input $I$ yields $\hat I$.

\subsection{Training Objectives}
\label{sec:training}

\noindent\textbf{Degradation axis alignment.}
For each category, the adapter $g$ maps its text embedding offset to a visual axis $\mathbf{a}_{\mathrm{deg}}$ (Sec.~\ref{sec:axes}). The loss $\mathcal{L}_{\mathrm{align}}$ maximizes cosine similarity between this axis and the window residual $\mathbf{r}_{\mathrm{deg}}^w$ from a matched clean and single-degradation pair, aligning the probe in Eq.~\eqref{eq:response} with the degradation-induced feature change.

\noindent\textbf{Coordinate calibration.}
We calibrate the shared coordinate head $h$ (Eq.~\eqref{eq:coordinates}) with a weighted sum of $\mathcal{L}_{\mathrm{zero}}$ and $\mathcal{L}_{\mathrm{rank}}$. The former anchors all coordinates at zero by averaging $|\tau_{\mathrm{deg}}^w(I_{\mathrm{clean}})|$ over clean images, categories, and windows.

For a clean image $I_{\mathrm{clean}}$, its single-degradation counterpart $I_{\mathrm{deg}}$, and a version $I_{\mathrm{deg}+}$ with a further increment of the same degradation, the successive changes in the selected coordinate are
\begin{equation}
\begin{aligned}
\delta\tau_{\mathrm{deg},0}^{w}
&= \tau_{\mathrm{deg}}^{w}(I_{\mathrm{deg}})
 - \tau_{\mathrm{deg}}^{w}(I_{\mathrm{clean}}),\\
\delta\tau_{\mathrm{deg},+}^{w}
&= \tau_{\mathrm{deg}}^{w}(I_{\mathrm{deg}+})
 - \tau_{\mathrm{deg}}^{w}(I_{\mathrm{deg}}).
\end{aligned}
\label{eq:ranking_increments}
\end{equation}
The ranking loss~\cite{RankIQA} penalizes each change below a margin $m>0$,
\begin{equation}
\mathcal{L}_{\mathrm{rank}}
=\mathbb{E}\Big[
[m-\delta\tau_{\mathrm{deg},0}^{w}]_+
+[m-\delta\tau_{\mathrm{deg},+}^{w}]_+
\Big],
\label{eq:ranking_loss}
\end{equation}
where $[x]_+=\max(0,x)$ and the expectation is over triplets, categories, and windows.

\noindent\textbf{Restoration training.}
After calibration, we freeze the measurement branch and compute fixed single-degradation response references from training windows. We pretrain the unconditioned baseline, then fine-tune it jointly with $\phi$ and the modulation modules (Sec.~\ref{sec:conditioning}). Both stages use the baseline restoration losses $\mathcal{L}_{\mathrm{pix}}$ and $\mathcal{L}_{\mathrm{freq}}$.

%% file: sec/4_expriments.tex
\section{Experiments}
\label{sec:expriments}

\begin{table*}[t]
    \centering
    \definecolor{cddBest}{RGB}{255,251,228}
    \definecolor{cddSecond}{RGB}{234,241,252}
    \definecolor{cddThird}{RGB}{243,244,246}
    \caption{
        Results on CDD-11 (PSNR / SSIM).
        For UniLDiff, we compute the average from the published per-condition scores.
        L: low-light, H: haze, R: rain, S: snow.
        \textit{Avg} is the macro-average over all 11 degradation conditions.
        Row colors mark the overall
        \colorbox{cddBest}{best},
        \colorbox{cddSecond}{second-best}, and
        \colorbox{cddThird}{third-best} methods
        (ranked by \textit{Avg} PSNR).
    }
    \vspace{-0.5em}
    \label{tab:cdd11}
    \setlength{\tabcolsep}{3pt}
    \renewcommand{\arraystretch}{1.15}
    \resizebox{\textwidth}{!}{%
        \begin{tabular}{@{}l*{12}{c}@{}}
            \toprule
            & \multicolumn{4}{c}{\textbf{CDD11-Single}}
            & \multicolumn{5}{c}{\textbf{CDD11-Double}}
            & \multicolumn{2}{c}{\textbf{CDD11-Triple}}
            & \\

            \cmidrule(lr){2-5}
            \cmidrule(lr){6-10}
            \cmidrule(lr){11-12}

            \textbf{Methods}
            & Low (L) & Haze (H) & Rain (R) & Snow (S)
            & L+H & L+R & L+S & H+R & H+S
            & L+H+R & L+H+S
            & \textbf{Avg} \\
            \midrule

            WeatherDiff\cite{WeatherDiff}
            & 23.58 / .763
            & 21.99 / .904
            & 24.85 / .885
            & 24.80 / .888
            & 21.83 / .756
            & 22.69 / .730
            & 22.12 / .707
            & 21.25 / .868
            & 21.99 / .868
            & 21.23 / .716
            & 21.04 / .698
            & 22.49 / .799 \\
            
            AirNet\cite{AirNet}
            & 24.83 / .778
            & 24.21 / .951
            & 26.55 / .891
            & 26.79 / .919
            & 23.23 / .779
            & 22.82 / .710
            & 23.29 / .723
            & 22.21 / .868
            & 23.29 / .901
            & 21.80 / .708
            & 22.24 / .725
            & 23.75 / .814 \\

            PromptIR\cite{PromptIR}
            & 26.32 / .805
            & 26.10 / .969
            & 31.56 / .946
            & 31.53 / .960
            & 24.49 / .789
            & 25.05 / .771
            & 24.51 / .761
            & 24.54 / .924
            & 23.70 / .925
            & 23.74 / .752
            & 23.33 / .747
            & 25.90 / .850 \\

            WGWS-Net\cite{WGWSNet}
            & 24.39 / .774
            & 27.90 / .982
            & 33.15 / .964
            & 34.43 / .973
            & 24.27 / .800
            & 25.06 / .772
            & 24.60 / .765
            & 27.23 / .955
            & 27.65 / .960
            & 23.90 / .772
            & 23.97 / .771
            & 26.96 / .863 \\

            OneRestore\cite{OneRestore}
            & 26.48 / .826
            & 32.52 / .990
            & 33.40 / .964
            & 34.31 / .973
            & 25.79 / .822
            & 25.58 / .799
            & 25.19 / .789
            & 29.99 / .957
            & 30.21 / .964
            & 24.78 / .788
            & 24.90 / .791
            & 28.47 / .878 \\
            
            MoCE-IR\cite{MoCEIR}
            & 27.26 / .824
            & 32.66 / .990
            & 34.31 / .970
            & 35.91 / .980
            & 26.24 / .817
            & 26.25 / .800
            & 26.04 / .793
            & 29.93 / .964
            & 30.19 / .970
            & 25.41 / .789
            & 25.39 / .790
            & 29.05 / .881 \\

            UniLDiff\cite{UniLDiff}
            & \textbf{27.45} / .830
            & 32.87 / .981
            & 34.60 / .972
            & 36.10 / \textbf{.983}
            & 26.50 / .820
            & 26.40 / .808
            & 26.10 / .805
            & 30.10 / .963
            & 30.50 / .964
            & 25.80 / .794
            & 25.50 / .790
            & 29.27 / .883 \\
            
            R2R\cite{R2R}
            & 27.29 / \underline{.836}
            & 32.88 / .992
            & 35.12 / .971
            & 36.54 / \underline{.981}
            & 26.21 / .833
            & \underline{26.70} / .816
            & \underline{26.33} / .809
            & 30.41 / .966
            & 30.37 / .971
            & 25.47 / .804
            & 25.21 / .803
            & 29.32 / .889 \\

            \rowcolor{cddThird}
            MIRAGE\cite{MIRAGE}
            & \underline{27.41} / .833
            & 33.12 / .992
            & 34.66 / .971
            & 35.98 / \underline{.981}
            & \underline{26.55} / .828
            & 26.53 / .810
            & \underline{26.33} / .803
            & 30.32 / .965
            & 30.27 / .969
            & 25.59 / .801
            & 25.86 / .799
            & 29.33 / .887 \\
            
            \rowcolor{cddSecond}
            CART\cite{CART}
            & 27.05 / \underline{.836}
            & \underline{34.52} / \underline{.994}
            & \textbf{35.46} / \textbf{.976}
            & \textbf{37.54} / \textbf{.983}
            & 26.40 / \underline{.836}
            & 26.40 / \underline{.819}
            & \underline{26.33} / \underline{.816}
            & \underline{31.61} / \underline{.974}
            & \underline{32.28} / \underline{.978}
            & \underline{25.93} / \underline{.813}
            & \underline{25.96} / \underline{.814}
            & \underline{29.95} / \underline{.894} \\

            \midrule
            \rowcolor{cddBest}
            \textbf{PROBE (Ours)}
            & 27.29 / \textbf{.841}
            & \textbf{36.95} / \textbf{.998}
            & \underline{35.21} / \underline{.975}
            & \underline{36.91} / \textbf{.983}
            & \textbf{27.56} / \textbf{.846}
            & \textbf{26.95} / \textbf{.825}
            & \textbf{26.87} / \textbf{.819}
            & \textbf{33.23} / \textbf{.975}
            & \textbf{34.32} / \textbf{.979}
            & \textbf{26.62} / \textbf{.821}
            & \textbf{26.52} / \textbf{.818}
            & \textbf{30.77} / \textbf{.898} \\
            \bottomrule
        \end{tabular}%
    }
    \vspace{-1em}
\end{table*}

\begin{table}[t]
    \centering
    \caption{Quantitative comparison on composite-degradation datasets. Bold values indicate the best reported results in each column.}
    \vspace{-0.5em}
    \label{tab:composite-comparison}

    \setlength{\tabcolsep}{4pt}
    \renewcommand{\arraystretch}{1.12}

    \resizebox{\columnwidth}{!}{%
        \begin{tabular}{@{}l*{6}{c}@{}}
            \toprule
            \multirow{2}{*}{Method}
            & \multicolumn{2}{c}{CDD-11~\cite{OneRestore}}
            & \multicolumn{2}{c}{AWMM-100K~\cite{AWMM}}
            & \multicolumn{2}{c}{SCD-15} \\
            \cmidrule(lr){2-3}
            \cmidrule(lr){4-5}
            \cmidrule(lr){6-7}
            & PSNR$\uparrow$ & SSIM$\uparrow$
            & PSNR$\uparrow$ & SSIM$\uparrow$
            & PSNR$\uparrow$ & SSIM$\uparrow$ \\
            \midrule
            AirNet~\cite{AirNet}
            & 23.75 & .814 & 27.23 & .847 & 29.42 & .889 \\
            OneRestore~\cite{OneRestore}
            & 28.47 & .878 & 29.71 & .896 & 30.43 & .901 \\
            PromptIR~\cite{PromptIR}
            & 25.90 & .850 & 32.35 & .918
            & 33.01 & .920 \\
            MoCE-IR~\cite{MoCEIR}
            & 29.05 & .881 & 32.90 & .924
            & 34.43 & .938 \\
            R2R~\cite{R2R}
            & 29.32 & .889
            & 32.70 & .923 & 32.28 & .918 \\
            \midrule
            \rowcolor{blue!7}
            \textbf{PROBE (Ours)}
            & \textbf{30.77} & \textbf{.898}
            & \textbf{34.90} & \textbf{.926} & \textbf{37.27} & \textbf{.939} \\
            \bottomrule
        \end{tabular}%
    }
    \vspace{-1em}
\end{table}

\subsection{Experimental Settings}

\noindent\textbf{Datasets.}
CDD-11~\cite{OneRestore} covers 11 single and composite conditions across low light, haze, rain, and snow. We evaluate AWMM-100K~\cite{AWMM} on paired degraded/clean visible-light images of rain, haze, snow, and their pairwise combinations.

SCD-15 uses 2,000 clean Cityscapes images~\cite{Cityscapes} (1,600 train, 400 test). The 15 nonempty combinations at three severities yield 45 variants per image. We train on the 11 CDD-11 conditions (52,800 pairs) and test on all 15 (18,000 pairs), holding out rain--snow, low-light--rain--snow, haze--rain--snow, and the four-factor combination from training. Synthesis, evaluation details, and examples appear in the supplementary material.

\noindent\textbf{Implementation details.}
We train PROBE in PyTorch on eight NVIDIA GeForce RTX 4090 GPUs with Adam, cosine decay from $2\times10^{-4}$, and the objectives in Sec.~\ref{sec:training}. The backbone uses $[1,1,1,28]$ encoder NAFBlocks, one bottleneck block, and one block per decoder stage. For composites, the measurement branch maps 64-dimensional features to $K$ coordinates per $16\times16$ window, where $K$ is the number of single-degradation types in the dataset. After calibration, it stays frozen during reference construction and restoration training. In the supplementary five-task evaluation, each input has one degradation, so we use coordinate conditioning alone, without bidirectional probing or response descriptors.

\begin{table}[t]
    \centering
    \caption{Complexity measured using 2 NVIDIA RTX 3090 GPUs with $224\times224$ patches. $^{*}$ denotes patch size of $64\times64$. ``--'' indicates values not reported in the original paper.}
    \label{tab:complexity}
    \vspace{-0.5em}
    \footnotesize
    \setlength{\tabcolsep}{1.5pt}
    \renewcommand{\arraystretch}{1.12}

    \begin{tabularx}{\columnwidth}{
        @{}l*{4}{>{\centering\arraybackslash}X}@{}
    }
        \toprule
        Method
        & \makecell{PSNR\\(dB)$\uparrow$}
        & \makecell{Params\\(M)$\downarrow$}
        & \makecell{Memory\\(MiB)$\downarrow$}
        & \makecell{MACs\\(G)$\downarrow$} \\
        \midrule
        AirNet~\cite{AirNet}
        & 23.75 & \textbf{8.93} & 255 & 231 \\
        PromptIR~\cite{PromptIR}
        & 25.90 & 35.59 & 584 & 133 \\
        WGWS-Net~\cite{WGWSNet}
        & 26.96 & 24.41 & 193 & 174 \\
        DOD~\cite{DOD}
        & -- & 1670.25 & 4613 & 950 \\
        WeatherDiff$^{*}$~\cite{WeatherDiff}
        & 22.49 & 82.96 & 2383 & 742 \\
        MoCE-IR~\cite{MoCEIR}
        & 29.05 & 25.35 & \underline{163} & 34 \\
        R2R~\cite{R2R}
        & \underline{29.32} & 30.59 & 225 & \textbf{13} \\
        \midrule
        \rowcolor{blue!7}
        \textbf{PROBE (Ours)}
        & \textbf{30.77} & \underline{17.73} & \textbf{128} & \underline{27} \\
        \bottomrule
    \end{tabularx}
    \vspace{-1em}
\end{table}

\subsection{Comparison with State-of-the-Art Methods}
\noindent\textbf{Composite degradations.}
On CDD-11, PROBE averages 30.77~dB/0.898 SSIM, exceeding CART by 0.82~dB/0.004 (Table~\ref{tab:cdd11}). It leads both metrics on all seven composite conditions, with the largest PSNR margins on haze--snow (2.04~dB) and haze--rain (1.62~dB). Single-condition PSNR leaders are PROBE for haze, UniLDiff for low light, and CART for rain and snow. UniLDiff's 11 published scores average 29.27~dB/0.883 versus the reported 29.35~dB/0.886~\cite{UniLDiff}.

PROBE exceeds MoCE-IR by 2.00~dB/0.002 SSIM on AWMM-100K and 2.84~dB/0.001 on SCD-15 at the reported precision (Table~\ref{tab:composite-comparison}). On AWMM-100K, PROBE scores 33.65~dB on haze--rain and 33.58~dB on haze--snow, while MoCE-IR leads on rain--snow (36.50 vs.\ 35.49~dB). PROBE leads all four unseen SCD-15 combinations in PSNR, averaging 35.37 vs.\ 33.39~dB for MoCE-IR. Condition-wise results appear in the supplementary material.

\input{sec/cdd_qualitative}

In both CDD-11 scenes (Fig.~\ref{fig:cdd-qualitative}), MoCE-IR~\cite{MoCEIR}, OneRestore~\cite{OneRestore}, and R2R~\cite{R2R} recover visibility but leave snow in the highlighted skies. PROBE better matches the clean skies, reducing veiling around distant buildings in haze--snow and restoring brightness and contrast while removing prominent snow in low-light--haze--snow. We analyze real-world composite restoration qualitatively in the supplementary material.

\begin{figure}[t]
    \centering
    \includegraphics[width=\linewidth]{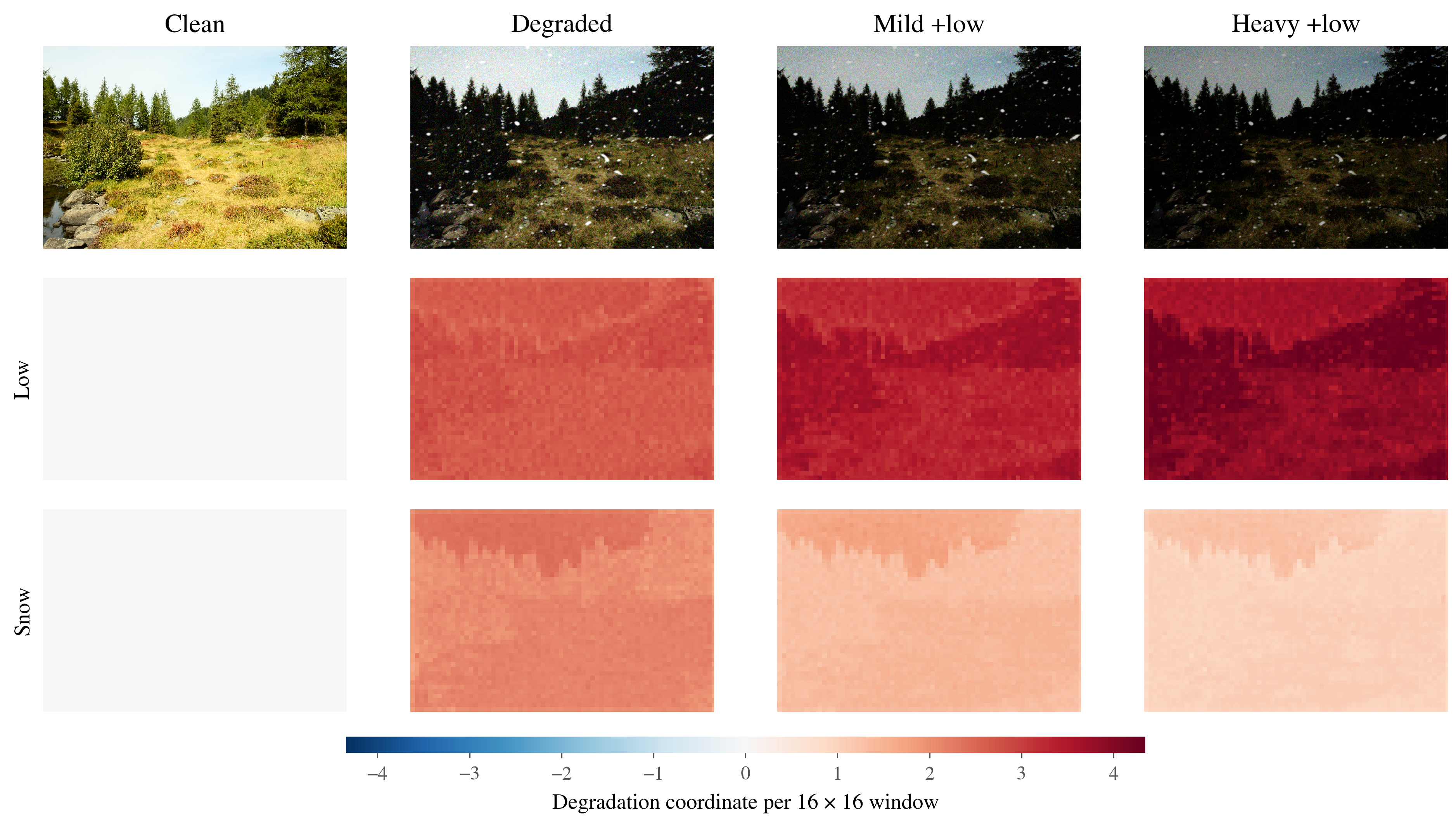}
    \vspace{-1.5em}
    \caption{To test how darkening affects the snow coordinate, we progressively darken a low-light--snow image with snow fixed, re-encoding and recomputing all four window-level coordinates at each step. On the shared scale, the low-light coordinate rises and the snow coordinate falls.}
    \label{fig:coordinate_low_snow}
    \vspace{-1em}
\end{figure}

\noindent\textbf{Complexity analysis.} 
PROBE achieves the highest CDD-11 PSNR (30.77~dB) and lowest memory use (128~MiB) in Table~\ref{tab:complexity}. Only AirNet has fewer parameters (8.93 vs.\ 17.73~M) and only R2R fewer MACs (13 vs.\ 27~G). We profile resources locally, using a smaller patch for WeatherDiff, and cite competitors' original PSNR values.

\begin{table}[t]
    \centering
    \small
    \caption{Core-component ablations on CDD-11.
    SDA: semantically anchored degradation axes;
    CRP: cross-degradation response probing.
    SDA-only uses degradation-category conditioning;
    CRP-only uses fixed random axes.
    Unused conditioning inputs are zeroed.
    Coordinates and responses are still computed internally
    when required to obtain $\Delta\mathbf{D}$.}
    \vspace{-0.5em}
    \label{tab:ablation_components}
    \setlength{\tabcolsep}{0pt}
    \renewcommand{\arraystretch}{1.15}

    \begin{tabular}{@{}
        *{2}{>{\centering\arraybackslash}p{0.12\columnwidth}}
        |
        *{3}{>{\centering\arraybackslash}p{0.12\columnwidth}}
        |
        *{2}{>{\centering\arraybackslash}p{0.20\columnwidth}}
        @{}}
        \toprule
        SDA & CRP
        & $\boldsymbol{\tau}$
        & $\mathbf{D}$
        & $\Delta\mathbf{D}$
        & PSNR$\uparrow$
        & SSIM$\uparrow$ \\
        \midrule

        \ding{55} & \ding{55}
        & \ding{55} & \ding{55} & \ding{55}
        & 28.98 & .884 \\

        \ding{51} & \ding{55}
        & \ding{55} & \ding{55} & \ding{55}
        & 29.44 & .888 \\

        \ding{55} & \ding{51}
        & \ding{51} & \ding{51} & \ding{51}
        & 28.63 & .883 \\

        \midrule

        \ding{51} & \ding{51}
        & \ding{51} & \ding{55} & \ding{55}
        & 30.17 & .892 \\

        \ding{51} & \ding{51}
        & \ding{55} & \ding{51} & \ding{55}
        & 29.76
        & .890 \\

        \ding{51} & \ding{51}
        & \ding{55} & \ding{55} & \ding{51}
        & 30.06
        & .891 \\

        \midrule

        \ding{51} & \ding{51}
        & \ding{51} & \ding{51} & \ding{55}
        & 30.38 & .894 \\

        \ding{51} & \ding{51}
        & \ding{51} & \ding{55} & \ding{51}
        & 30.63 & .896 \\

        \ding{51} & \ding{51}
        & \ding{55} & \ding{51} & \ding{51}
        & 30.10
        & .892 \\

        \midrule

        \ding{51} & \ding{51}
        & \ding{51} & \ding{51} & \ding{51}
        & \textbf{30.77} & \textbf{.898} \\

        \bottomrule
    \end{tabular}
    \vspace{-1em}
\end{table}

\subsection{Ablation Study}
Tables~\ref{tab:ablation_components} and~\ref{tab:ablation_design} evaluate axis construction (SDA) and response probing (CRP) under a common training protocol. Scores are macro-averaged over 11 CDD-11 conditions.

\noindent\textbf{Module contributions.} Category-conditioned SDA alone, with all descriptors disabled, raises backbone PSNR from 28.98 to 29.44~dB. Random axes with all descriptors yield 28.63~dB, 0.35~dB below baseline. With both modules enabled, coordinates lead among individual inputs (30.17~dB), followed by reference-relative deviations (30.06~dB) and normalized responses (29.76~dB).

Adding normalized responses or reference-relative deviations to coordinates improves PSNR by 0.21 or 0.46~dB, respectively. All three reach 30.77~dB/0.898 SSIM, 0.14~dB above coordinates plus deviations. Omitting coordinates from the descriptor yields 30.10~dB, although computing deviations still requires them.

\begin{table}[t]
    \centering
    \small
    \caption{Design ablations on CDD-11. Each variant changes the indicated design relative to PROBE (full). Best results are bold, including ties at the reported precision.}
    \vspace{-0.5em}
    \label{tab:ablation_design}
    \setlength{\tabcolsep}{4pt}
    \setlength{\aboverulesep}{0.35ex}
    \setlength{\belowrulesep}{0.45ex}
    \renewcommand{\arraystretch}{1.0}

    \begin{tabularx}{\columnwidth}{@{}>{\raggedright\arraybackslash}X cc@{}}
        \toprule
        Configuration & PSNR$\uparrow$ & SSIM$\uparrow$ \\
        \midrule

        \multicolumn{3}{@{}l}{\textit{(a) Axis construction}} \\
        Random axes & 28.63 & .883 \\
        Learned visual-only axes & 30.45 & .894 \\

        \midrule
        \multicolumn{3}{@{}l}{\textit{(b) Restoration conditioning}} \\
        Bottleneck-only injection & 30.47 & .894 \\
        Decoder-only injection & 30.60 & .895 \\

        \midrule
        \textbf{PROBE (full)} & \textbf{30.77} & \textbf{.898} \\
        \bottomrule
    \end{tabularx}
    \vspace{-1em}
\end{table}

\noindent\textbf{Axis construction.} With all descriptors, learned visual-only axes improve on random axes (30.45 vs.\ 28.63~dB). Semantic anchoring adds 0.32~dB (Table~\ref{tab:ablation_design}).

\noindent\textbf{Restoration conditioning.}
Joint conditioning reaches 30.77~dB versus 30.47 for bottleneck-only and 30.60 for decoder-only injection (Table~\ref{tab:ablation_design}).

\noindent\textbf{Predicting coordinate changes under darkening.} On 200 held-out CDD-11 low-light--snow scenes, we compare per-window feature-space predictions (low-light-to-snow, Eq.~\eqref{eq:response}) with snow-coordinate changes after gamma darkening and re-encoding. The gamma-based darkening parameter $l$ differs from the synthesis strength $d$ in Sec.~\ref{sec:analysis}. Mild ($l=0.65$) and strong ($l=1.20$) darkening yield 92.71\% and 93.35\% sign agreement and mean within-scene Spearman correlations of 0.537 and 0.564, respectively. In Fig.~\ref{fig:coordinate_low_snow}, both coordinates are near zero for the clean image and positive for the composite. Further darkening with snow fixed raises the low-light coordinate and lowers the snow coordinate, complementing Fig.~\ref{fig:motivation2}'s pixel responses to a fixed snow-opacity increment at different darkening levels.

\noindent\textbf{Cross-degradation responses without direct supervision.} Eq.~\eqref{eq:ranking_loss} supervises only the enhanced coordinate. A scan of probe directions on 200 CDD-11 low-light--snow test scenes identified the rain-to-snow response. We compared per-window predictions from Eq.~\eqref{eq:response} with snow-coordinate changes after adding rain (strength 0.05) and re-encoding. Mean within-scene sign agreement was 99.2\%, and within-scene Spearman correlations were positive in 196/200 scenes (98\%).

\begin{figure}[t]
    \centering
    \includegraphics[width=0.8\columnwidth]{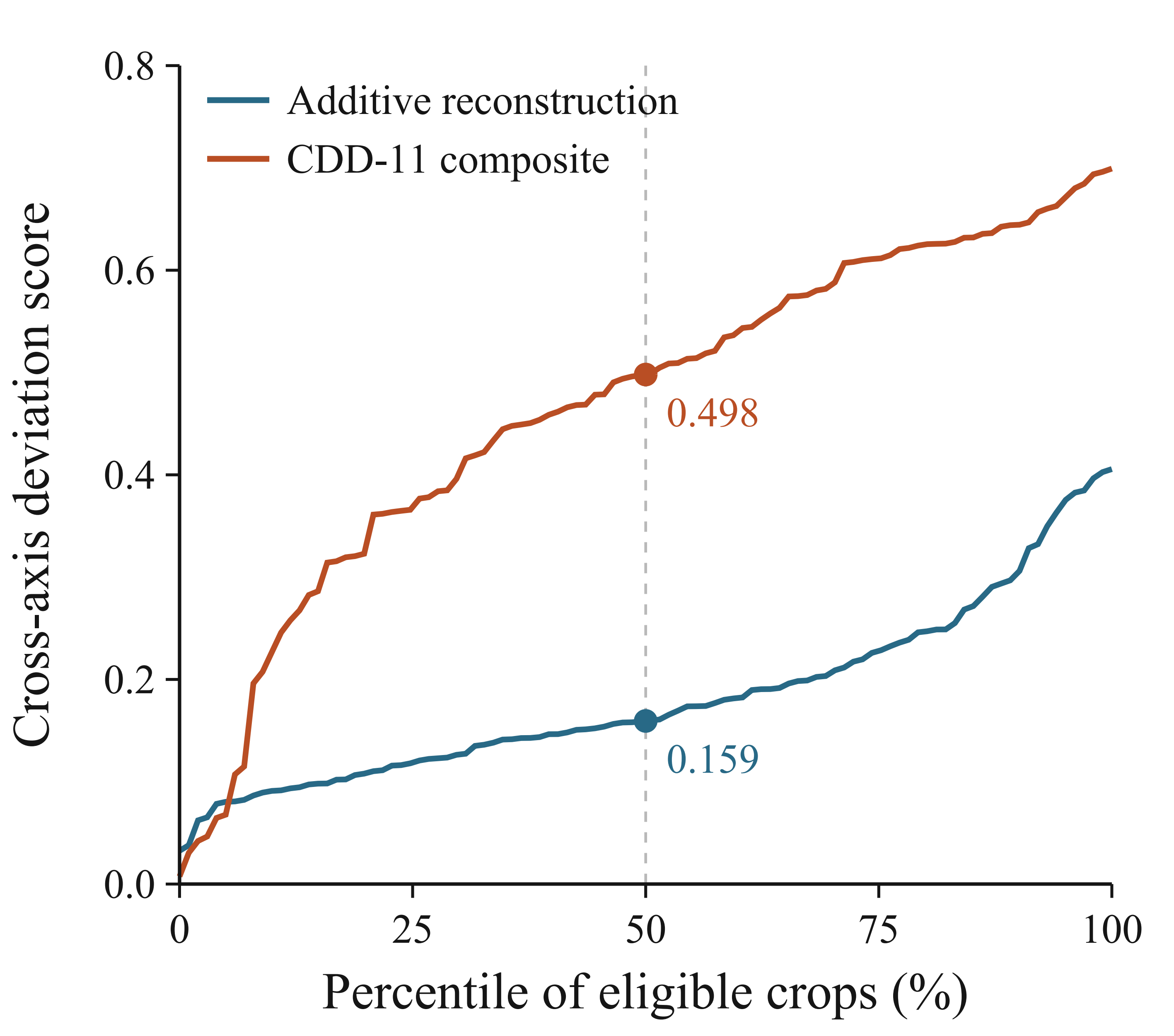}
    \vspace{-0.5em}
    \caption{To test for larger cross-axis deviations in low-light--haze composites, we score 102 matched CDD-11 crop pairs and plot $S_{\mathrm{LH}}$ (Eq.~\eqref{eq:lh_crop_score}) by within-group percentile. The dashed 50th-percentile line marks the additive and composite medians (0.159 and 0.498).}
    \label{fig:paired_additive_percentiles}
    \vspace{-1em}
\end{figure}

\noindent\textbf{Characterizing non-additive responses with paired controls.}
To test for larger cross-axis deviations in composites, we compare aligned CDD-11 crops from $I_{\mathrm{LH}}$ and its pixel-additive control $I_{\mathrm{add}}=I_{\mathrm{L}}+I_{\mathrm{H}}-I_{\mathrm{clean}}$, constructed from same-scene low-light, haze, and clean images. In window $w$, $\Delta\mathbf{D}^w$ subtracts the state-weighted single-degradation reference queried at $\boldsymbol{\tau}^w$ from the scale-normalized response (Eq.~\eqref{eq:reference}). For crop $I$, $S_{\mathrm{LH}}(I)$ is the RMS of normalized low-light-to-haze and haze-to-low-light deviations over its $N$ windows:
\begin{equation}
S_{\mathrm{LH}}(I)=
\sqrt{\frac{1}{2N}\sum_{w=1}^{N}
\left[(\Delta D^w_{\mathrm{L}\mathrm{H}})^2+
      (\Delta D^w_{\mathrm{H}\mathrm{L}})^2\right]}.
\label{eq:lh_crop_score}
\end{equation}
Each crop is scored separately. Larger scores indicate greater deviation in either direction, regardless of sign.
Of 200 candidate crops, 102 have additive controls within the valid RGB range without clipping.

Figure~\ref{fig:paired_additive_percentiles} shows higher cross-axis deviations for composites across the central percentiles. Within matched pairs, composites score higher than their additive controls in 93 of 102 cases (91.2\%), with a mean paired difference of 0.285. The paired scatter plot appears in the supplementary material.

%% file: sec/cdd_qualitative.tex
\begin{figure*}[t]
  \centering
  \begingroup
  \setlength{\tabcolsep}{0pt}
  \newcommand{\cddwidth}{\dimexpr\linewidth-1.6em\relax}
  \newcommand{\cddhead}[1]{\makebox[\dimexpr(\cddwidth)/6\relax][c]{\small #1}}
  \newcommand{\cddrow}[2]{%
    \makebox[1.6em][c]{\raisebox{-.5\height}{\rotatebox{90}{\footnotesize #1}}}&%
    \raisebox{-.5\height}{\includegraphics[width=\cddwidth]{sec/figures/qualitative_vis/#2}}}
  \begin{tabular}{@{}cc@{}}
    &\cddhead{Degraded}\cddhead{MoCE-IR}\cddhead{OneRestore}\cddhead{R2R}\cddhead{PROBE (Ours)}\cddhead{GT}\\[1pt]
    \cddrow{Haze--Snow}{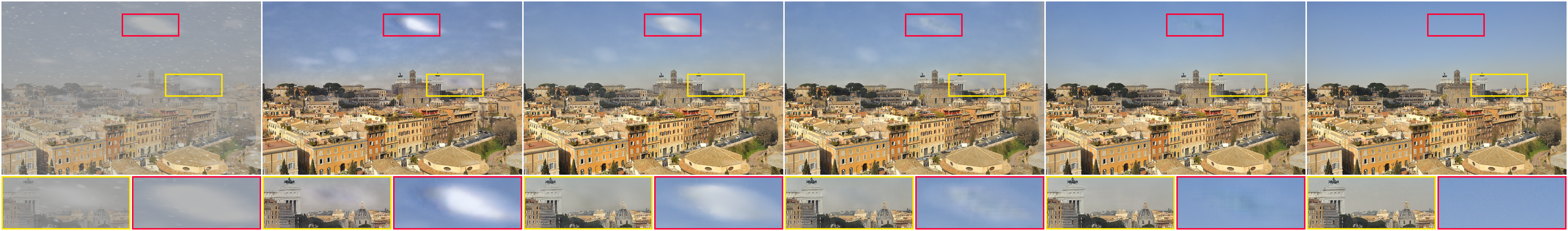}\\[1pt]
    \cddrow{Low--Haze--Snow}{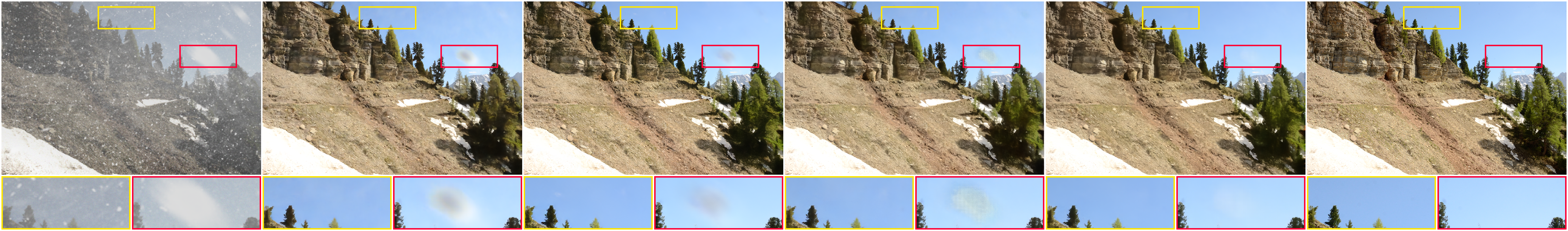}
  \end{tabular}
  \endgroup
  \vspace{-0.5em}
  \caption{Qualitative comparison on CDD-11 under haze--snow and low-light--haze--snow degradations. Below each image, the left and right crops magnify the yellow and red regions, respectively.}
  \label{fig:cdd-qualitative}
  \vspace{-1em}
\end{figure*}

%% file: sec/5_conclusion.tex
\section{Conclusion}
\label{sec:conclusion}

We introduce PROBE, an interventional framework that shifts the focus of composite image restoration from static degradation-state estimation to dynamic, state-dependent response modeling. Semantically anchored, visually calibrated axes define explicit perturbation directions for measuring local cross-coordinate responses. Comparing these responses with intensity-matched single-degradation references explicitly quantifies non-additive coupling overlooked by prior factor-wise approaches. The resulting descriptors and calibrated degradation coordinates provide a principled, adaptive basis for high-fidelity reconstruction. To comprehensively validate PROBE across diverse composite degradation conditions, we introduce SCD-15 to assess sensitivity to degradation severity and generalization to combinations held out during restoration training. Across extensive experiments, PROBE consistently achieves state-of-the-art performance on CDD-11, AWMM-100K, and SCD-15. Our findings underscore that modeling local inter-factor dependencies is not merely a refinement, but a fundamental requirement for mastering the complex, non-additive nature of real-world composite degradations.

%% file: sec/LLM_Usage_Statement.tex
\section*{LLM Usage Statement}
We used OpenAI Codex to assist with language polishing,
code debugging, and manuscript consistency checks.
The authors reviewed its suggestions and take responsibility
for the final manuscript, implementation, and reported results.